# Towards Practical Multimodal Hospital Outbreak Detection

Chang Liu, MS[1], Jieshi Chen, MS[1], Alexander J. Sundermann, DrPH[2,3,4], Kathleen Shutt, MS [2,3], Marissa P. Griffith, BS [2,3], Lora Lee Pless, PhD [2,3],
Lee H. Harrison, MD[2,3,4], Artur W. Dubrawski, PhD[1].

[1]Auton Lab, Carnegie Mellon University, Pittsburgh, Pennsylvania, USA;
[2]Microbial Genomic Epidemiology Laboratory, Center for Genomic Epidemiology, University of Pittsburgh, Pittsburgh, Pennsylvania, USA;
[3]Division of Infectious Diseases, University of Pittsburgh School of Medicine, Pittsburgh, Pennsylvania, USA;
[4]Department of Epidemiology, School of Public Health, University of Pittsburgh, Pittsburgh, Pennsylvania, USA.

**Abstract**
*Rapid identification of outbreaks in hospitals is essential for controlling pathogens with epidemic potential. Although whole genome sequencing (WGS) remains the gold standard in outbreak investigations, its substantial costs and turnaround times limit its feasibility for routine surveillance, especially in less-equipped facilities. We explore three modalities as rapid alternatives: matrix-assisted laser desorption ionization-time of flight (MALDI-TOF) mass spectrometry, antimicrobial resistance (AR) patterns, and electronic health records (EHR). We present a machine learning approach that learns discriminative features from these modalities to support outbreak detection. Multi-species evaluation shows that the integration of these modalities can boost outbreak detection performance. We also propose a tiered surveillance paradigm that can reduce the need for WGS through these alternative modalities. Further analysis of EHR information identifies potentially high-risk contamination routes linked to specific clinical procedures, notably those involving invasive equipment and high-frequency workflows, providing infection prevention teams with actionable targets for proactive risk mitigation.*

**Introduction**
Hospital-acquired infections can lead to outbreaks when pathogenic organisms transmit uncontrollably between patients[1], healthcare workers[2], and equipment[3]. Early detection and contention of these outbreaks are critical to minimizing patient morbidity and mortality and to reducing the institutional burden required to stem large-scale transmissions[4]. To address this need, infection prevention teams in healthcare facilities gather epidemiological information from electronic health record (EHR) systems alongside microbiological data from patient and environmental specimens to identify potential outbreak clusters[5]. These clusters are defined by shared characteristics of involved pathogens, indicating their transmission routes[6].

Providing the richest information and highest resolution, whole genome sequencing (WGS) stands as the gold standard for determining pathogen similarity[7] and is adopted to better understand pathogen transmission dynamics[8]. The methodology relies on quantifying genetic differences between isolates through single-nucleotide polymorphisms (SNPs). Cases are assigned to the same outbreak cluster when their SNP distance falls below a predetermined cutoff[9]. However, substantial expenses, extended processing times, and specialized expertise requirements have limited the routine implementation of WGS in clinical laboratories[10].

To enable rapid and accessible outbreak detection, we investigated three alternative modalities: matrix-assisted laser desorption ionization-time of flight (MALDI-TOF) mass spectrometry, antimicrobial resistance (AR) patterns, and electronic health records (EHR). MALDI-TOF is a standard tool for species identification in clinical microbiology[11] and has been proposed as a promising WGS alternative for outbreak cluster detection given its low cost and rapid processing time[12]. It has demonstrated utility in epidemiological investigations of multiple pathogen species[13].

Analysis of AR profiles for outbreak detection presents an appealing approach by utilizing data routinely generated in clinical laboratories, thereby enabling near instantaneous analysis[14] and requiring minimal resources[15]. AR has proved effective in identifying outbreak clusters involving *Shigella* spp.[15] and VRE[16], with the potential to synergize with MALDI-TOF to enhance outbreak detection performance[17].

EHR data provides continuous epidemiological context captured during routine clinical care. EHR has demonstrated utility in outbreak surveillance through expediting contact tracing[18] and automating transmission route inference[19]. Additionally, EHR information has been shown to synergize with WGS by either jointly inferring outbreak clusters[19] or assigning unsequenced isolates to existing clusters defined by WGS[20]. While these applications establish the value of EHR for infection control response, they do not explore EHR's potential for directly assessing pathogen similarity in outbreak detection, nor its synergistic capacity with other non-WGS modalities such as MALDI-TOF or AR.

To address these gaps, we develop a machine learning framework that learns informative representations from MALDI-TOF spectra, AR patterns, and EHR features for hospital outbreak detection. We perform multi-species analyses comparing the performance of these three modalities, individually and in combination, against WGS-based ground truth across both species-agnostic and species-specific scenarios. Our findings demonstrate that integrating MALDI-TOF, AR, and EHR data can significantly enhance outbreak detection performance. Furthermore, we propose a tiered surveillance paradigm that utilizes these alternative modalities as front-line triage tools to reduce WGS burden. Finally, by interrogating the EHR features driving cluster detection, we identify high-risk clinical procedures primarily related to equipment contamination and workflow exposures. This capability pinpoints specific procedural vulnerabilities, allowing infection prevention teams to more effectively target their proactive risk mitigation activity.

**Methods**
*Data overview*: Our study utilized a proprietary and de-identified dataset of 4,921 isolates across 17 bacterial species (Table 1), collected during routine surveillance from patients at a large non-profit academic hospital between October 2021 and October 2024 to track active hospital outbreaks. Each isolate was associated with a MALDI-TOF spectrum, a list of phenotypic antimicrobial resistance patterns, and EHR data indicating the historical clinical procedures the patient had undergone.

To establish the ground truth outbreak clusters formed among these isolates, we leveraged pairwise SNP distances derived from WGS of isolates within the same species. Ground truth outbreak clusters were defined through complete-linkage hierarchical clustering on the SNP distance matrices, using a cutoff distance of 15. This threshold mirrors the institutional standard utilized by the hospital's infection prevention team for outbreak detection.

We evaluated our methods using cluster-grouped 4-fold cross-validation, stratified by species labels. This splitting strategy preserved the species distribution across data folds while ensuring all isolates from a single ground-truth cluster remained in the same fold. To ensure rigorous evaluation and prevent data leakage, only the SNP distances among isolates within the training folds were visible to the model during cross-validation.

*Modeling MALDI-TOF spectra and AR patterns:* To extract feature representations for MALDI-TOF and AR data, we utilized the data processing and encoding method established and validated in our previous research[17]. MALDI-TOF data consists of paired mass-to-charge ($m/z$) ratios and intensities. We retained the pairs with $m/z$ ratios between 2000 to 20000 Daltons and discretized the resulting spectra into 2000 uniform bins. To encode these spectra, we employed a 1D convolutional neural network, which projects the inputs onto a compact, 128-dimensional embedding space. The network was optimized using a dual-objective loss function that simultaneously predicts the isolate species and maps the Euclidean distances of the learned embeddings to their ground truth SNP distances (Figure 1a). Complete architectural specifications and loss formulations are detailed in our previous work[17].

The encoding strategy for AR data remains consistent with our prior methodology[17]. Specifically, antimicrobial resistance outcomes across a panel of 71 antibiotics were deterministically mapped into 2-dimensional vectors: "Sensitive" → [1,0], "Intermediate" → [0.5,0.5], "Resistant" → [0,1] (Figure 1b). Unmeasured tests were explicitly encoded as [0,0] to represent maximum uncertainty orthogonally to the empirical outcomes. Concatenating these values yielded a flattened, 142-dimensional feature vector for each isolate, capturing both its specific resistance phenotype and its clinical testing availability.

*Modeling EHR data*: To capture the epidemiological and environmental context of each isolate, we extracted the historical clinical procedure codes documented during the patient's hospital stay, strictly limited to a 30-day window prior to the isolate's culture date. To encode these textual identifiers, we employed a two-stage strategy (Figure 1c): latent semantic analysis (LSA)[21] and neural network encoding.

**Table 1.** Species-specific dataset statistics. **# Clusters** represents the number of outbreak clusters with more than one isolate. **Max Cluster Size** represents the largest cluster per species. **% Clustered** denotes the percentage of isolates involved in outbreak clusters rather than occurring as sporadic singletons.

| Species | Abbr. | # Isolates | # Clusters | Max Cluster Size | % Clustered |
|---|---|---|---|---|---|
| *Acinetobacter baumannii* | ACIN | 137 | 8 | 15 | 23.4 |
| *Burkholderia cepaciae* | BC | 28 | 4 | 7 | 57.1 |
| *Citrobacter* | CB | 105 | 7 | 3 | 15.2 |
| *Enterobacter cloacae* | EB | 113 | 13 | 4 | 25.7 |
| *Escherichia coli* | EC | 303 | 26 | 4 | 20.8 |
| *Klebsiella oxytoca* | KLO | 45 | 2 | 3 | 11.1 |
| *Klebsiella pneumoniae* | KLP | 273 | 41 | 11 | 49.5 |
| *Staphylococcus aureus* | MRSA | 544 | 44 | 6 | 18.9 |
| *Mycobacterium* | MYC | 87 | 8 | 14 | 39.1 |
| *Proteus mirabilis* | PR | 683 | 53 | 7 | 19.6 |
| *Providencia* | PRV | 81 | 2 | 3 | 6.2 |
| *Pseudomonas aeruginosa* | PSA | 1543 | 176 | 46 | 36.0 |
| *Pseudomonas* | PSB | 66 | 2 | 2 | 6.1 |
| *Serratia marcescens* | SER | 364 | 40 | 8 | 29.9 |
| *Stenotrophomonas maltophilia* | STEN | 268 | 18 | 9 | 20.9 |
| *Vancomycin-resistant Enterococcus* | VRE | 270 | 22 | 22 | 49.3 |
| Multiple species (others) | MULT | 11 | 2 | 2 | 36.4 |
| Total | | 4921 | 468 | 46 | 29.1 |

During the LSA stage, patient procedural histories were vectorized using Term Frequency-Inverse Document Frequency (TF-IDF). We utilized a custom identity tokenizer that strictly preserved the original casing and exact alphanumeric formatting of the clinical identifiers. To isolate meaningful transmission signals from routine clinical noise, we applied empirical frequency boundaries: highly sporadic procedures (associated with fewer than 10 isolates) were excluded to prevent overfitting, while ubiquitous procedures (present in over 90% of patient records) were removed as non-discriminative baseline care. This filtering strategy yielded a sparse 2750-dimensional feature vector for each isolate. To reduce dimensionality and extract the core underlying clinical pathways, we applied truncated singular value decomposition (Truncated SVD) to the TF-IDF matrices, resulting in dense, 256-dimensional feature vectors that retained 70.12% of the original TF-IDF variance.

In the neural network encoding stage, these 256-dimensional feature vectors were passed through a multilayer perceptron (MLP) consisting of two fully connected layers with Sigmoid Linear Unit (SiLU)[22] activations. This network was trained to non-linearly remap clinical pathways into a 256-dimensional epidemiological embedding space. Mirroring the optimization strategy of the MALDI-TOF encoder, this MLP utilized a distance mapping objective that forced the Euclidean distances between the learned EHR embeddings to approximate their corresponding ground-truth pairwise SNP distances.

*Multimodal integration*: To leverage and assess the complementarity of different epidemiological signals, we integrated the unimodal representations into a unified feature vector. Following the procedure in our previous work[17], to ensure scale and dimensional compatibility, all feature vectors were first standardized to zero-mean and unit-variance. The standardized AR (142-dimensional) and EHR (256-dimensional) representations were then respectively projected onto 128-dimensional spaces using Principal Component Analysis (PCA) to align with the dimensionality of the MALDI-TOF features (Figure 1d).

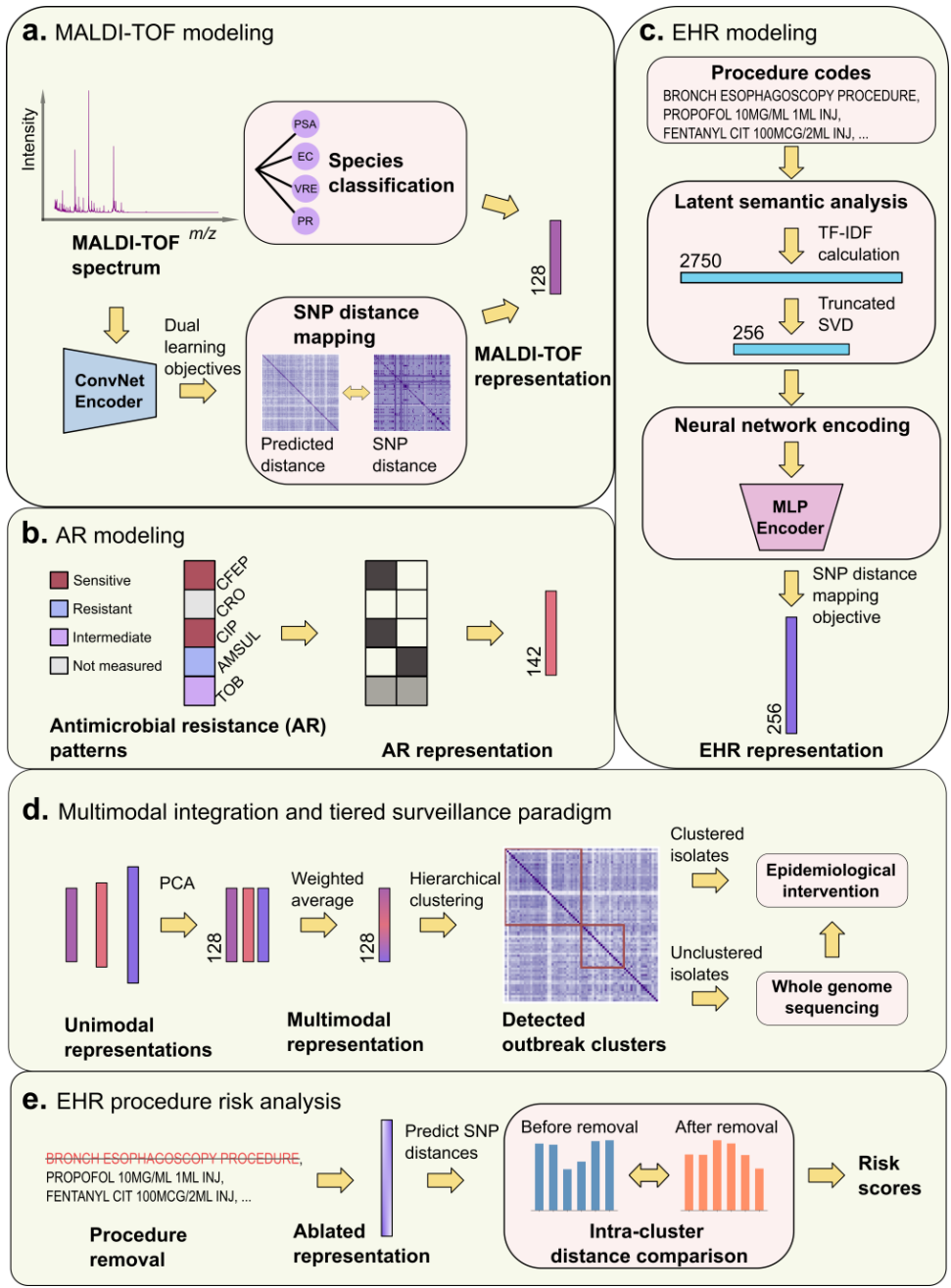

**Figure 1**. Framework overview. **a**. MALDI-TOF spectrum modeling via a dual-objective convolutional neural network. **b**. Antimicrobial resistance (AR) modeling encoding susceptibility phenotypes into low-dimensional features. **c**. Electronic Health Record (EHR) modeling utilizing latent semantic analysis and a neural network to predict SNP distances. **d**. Multimodal integration and tiered surveillance. Representations are combined via PCA and weighted averaging for hierarchical clustering to predict outbreak clusters. Unclustered isolates are triaged for whole-genome sequencing (WGS) confirmation. **e**. EHR procedure risk analysis. High-risk clinical pathways are identified via counterfactual feature ablation, comparing predicted intra-cluster SNP distances before and after procedure removal.

The aligned 128-dimensional vectors were subsequently fused via a weighted sum. The modality-specific coefficients for this summation were empirically optimized to maximize downstream clustering performance. For the trimodal framework (MALDI+AR+EHR), the optimal fusion weights were established as 0.2, 0.6, and 0.2, respectively. For bimodal fusion, the optimal weight distribution was 0.25 and 0.75 for the MALDI+AR configuration, and an equal 0.5-0.5 split for other pairwise combinations (MALDI+EHR and AR+EHR).

*Detecting and evaluating outbreak clusters*: To group the isolates into predicted outbreak clusters, the representations were subjected to agglomerative hierarchical clustering using Ward's linkage[23]. The optimal distance threshold for defining these clusters was determined independently for each species and cross-validation fold by maximizing the outbreak F1 score, the harmonic mean of outbreak sensitivity and positive predictive value (PPV), on the training data.

To evaluate clustering performance, we compared the resulting clusters against the WGS-derived ground truth clusters. Following our previously established evaluation framework[17], we evaluated the clustering assignments as a pairwise binary classification task, where the positive class denotes isolate pairs belonging to the same true transmission cluster. Performance was quantified using both standard machine learning and clinical epidemiological metrics. For unsupervised clustering fidelity, we reported the Normalized Mutual Information (NMI) and the Adjusted Rand Index (ARI). To assess clinical utility, we calculated outbreak sensitivity (the proportion of true outbreak clusters successfully identified) and PPV (the proportion of predicted clusters confirmed by WGS ground truth).

*Tiered surveillance paradigm*: We designed a tiered surveillance framework in which alternative modalities function as front-line triage tools rather than a replacement for WGS. Under this paradigm, isolates not assigned to a predicted cluster are referred to WGS confirmation (Figure 1d). To evaluate the utility of this framework, we calculated the percentage of isolates requiring WGS ($p$) as the proportion of predicted singleton isolates. We defined the "combined sensitivity" and "combined PPV" as the weighted average of the performance of the alternative modality (e.g., MALDI+EHR) and that of WGS. Given that WGS serves as the ground truth, its sensitivity and PPV are both defined as 1:

$$\text{Combined Sensitivity} = \text{Sensitivity}_{\text{model}} \times (1-p) + p,$$
$$\text{Combined PPV} = \text{PPV}_{\text{model}} \times (1-p) + p,$$

Where $\text{Sensitivity}_{\text{model}}$ and $\text{PPV}_{\text{model}}$ represent the sensitivity and PPV of the alternative modality, respectively. These combined metrics quantify the utility of different modalities in reducing WGS reliance while maintaining outbreak detection accuracy.

*Identifying potentially high-risk procedures*: To isolate the specific clinical pathways most indicative of shared environmental exposure, we performed a counterfactual feature ablation analysis on the EHR representations (Figure 1e). First, all clinical procedure codes within the dataset were manually grouped into 121 categories based on their clinical function. To quantify the epidemiological significance of a specific group, we employed a "leave-one-group-out" strategy: for each non-singleton isolate, we systematically removed all procedural identifiers belonging to a target group from the patient's record. The perturbed procedural histories were then re-processed through the frozen TF-IDF, Truncated SVD, and MLP encoding pipeline to generate modified neural embeddings.

The risk associated with each procedure group was quantified by the Wasserstein distance between the distribution of predicted SNP distances (relative to other members of the same cluster) before and after the procedures' artificial removal. A large Wasserstein distance indicates a significant drift in the isolate's latent representation, identifying the cluster's formation as highly sensitive to those clinical procedures. Note that to ensure the numeric stability of the Wasserstein distance, we restricted the calculation and comparison of the risk scores to those clusters with at least five isolates. Finally, we cross-referenced these risk scores with the cluster size and the intra-cluster prevalence of the group to ensure the identified high-risk procedures represented consistent transmission signals worthy of the infection control team's attention. We also recorded the change in the mean predicted SNP distance as additional reference. The metrics are mathematically defined as follows:

$$D_{\text{before}} = \{d(v,j) \mid j \in C \setminus \{v\}\}, \; D_{\text{after}} = \{d(\tilde{v},j) \mid j \in C \setminus \{v\}\},$$
$$\text{Risk} = \mathcal{W}_1(D_{\text{before}}, D_{\text{after}}) = \int_{-\infty}^{\infty} |F_{\text{before}}(x) - F_{\text{after}}(x)| \, dx,$$
$$\Delta \text{SNP}_{\text{pred}} = |\mu(D_{\text{after}}) - \mu(D_{\text{before}})|,$$

Here, $v$ is one isolate in cluster $C$ with the procedures of interest, and $\tilde{v}$ represents the same isolate after those procedures have been removed from its EHR feature. $d(\cdot,\cdot)$ denotes the predicted SNP distance between two specified isolates. $D_{\text{before}}$ and $D_{\text{after}}$ represent the distribution of predicted distances before and after the procedures' removal, respectively. $F_{\text{before}}$ and $F_{\text{after}}$ represent the cumulative distribution functions (CDFs) of these distance distributions, and $\mu(\cdot)$ denotes the mean of the specified distance distribution.

**Table 2.** Overall clustering performance (mean ± 95% confidence interval) of multimodal and unimodal alternatives to WGS. The best result for each metric is highlighted in **bold**, while the second best is marked in *italics*.

| Model | NMI | ARI | Sensitivity | PPV | F1 score |
|---|---|---|---|---|---|
| MALDI+AR+EHR | **0.831 ± 0.020** | *0.205 ± 0.014* | 0.190 ± 0.049 | **0.295 ± 0.057** | *0.212 ± 0.014* |
| MALDI+AR | 0.783 ± 0.019 | 0.165 ± 0.022 | 0.209 ± 0.041 | 0.188 ± 0.024 | 0.175 ± 0.021 |
| MALDI+EHR | *0.820 ± 0.017* | **0.205 ± 0.013** | 0.200 ± 0.052 | *0.272 ± 0.040* | **0.213 ± 0.013** |
| AR+EHR | 0.793 ± 0.027 | 0.118 ± 0.011 | 0.146 ± 0.003 | 0.125 ± 0.017 | 0.130 ± 0.010 |
| MALDI | 0.731 ± 0.017 | 0.160 ± 0.009 | *0.256 ± 0.059* | 0.143 ± 0.010 | 0.173 ± 0.009 |
| AR | 0.419 ± 0.006 | 0.009 ± 0.000 | **0.577 ± 0.003** | 0.018 ± 0.001 | 0.034 ± 0.001 |
| EHR | 0.760 ± 0.014 | 0.091 ± 0.004 | 0.170 ± 0.013 | 0.083 ± 0.001 | 0.106 ± 0.004 |

**Results**

*Distinct signals show synergistic potential in outbreak detection*: To evaluate the contributions of microbiological and clinical data to outbreak detection, we compared the clustering performance of all three modalities in unimodal, bimodal, and trimodal configurations across the full 17-species dataset. All performance metrics were calculated exclusively for isolates belonging to ground truth clusters of at least two members to prevent distortion from sporadic singleton cases. The values reported in Table 2 represent the mean and 95% confidence interval (using a $t$-distribution) across five independent trials, each utilizing a unique random seed for 4-fold cross-validation and model initialization.

We first observed distinct performance trade-offs among the unimodal methods. Specifically, MALDI-TOF emerged as the most robust single modality for cluster precision, yielding superior ARI (0.160), PPV (0.143), and F1 score (0.173). In contrast, AR and EHR demonstrated their own unique strengths, achieving the highest sensitivity (0.577) and NMI (0.760), respectively. These discrepancies suggest that the three modalities capture fundamentally distinct epidemiological features of pathogen transmission.

The integration of these modalities consistently enhanced performance. All three bimodal configurations (MALDI+AR, MALDI+EHR, AR+EHR) exhibited across-the-board gains in NMI, ARI, PPV, and F1 score compared to their unimodal components. This suggests that the feature spaces of these modalities are highly complementary, offering effective synergy for outbreak detection. Notably, the MALDI+EHR configuration achieved the highest overall outbreak F1 score (0.213). Trimodal integration (MALDI+AR+EHR) further extended these gains and attained the highest overall performance in NMI (0.831) and PPV (0.295), providing the most comprehensive reconstruction of complex transmission clusters.

*Multimodal integration enhances utility across diverse pathogens*: To characterize how multimodal integration improves outbreak detection across varying biological contexts, we performed a pathogen-specific analysis of sensitivity and PPV. We examined how the species-specific utility of MALDI-TOF—the single modality with the highest overall F1 score—is influenced by the inclusion of EHR and AR data. As illustrated in Figure 2a, integrating additional modalities with MALDI-TOF enhanced both sensitivity and PPV for the majority of species, though the optimal modality combination varied substantially by pathogen.

For instance, pathogens such as CB, EC, KLO, MRSA, and PR demonstrated broad improvements across all multimodal models. In other cases, gains were more modality-specific: adding AR data alone provided the greatest benefit for BC, whereas EHR integration proved most effective for EB, KLO, and PSA. In contrast, integration did not yield measurable gains for ACIN, MYC, or PRV. Furthermore, certain pathogens exhibited stability in specific metrics regardless of the model used: sensitivity showed only marginal fluctuations for MRSA, PR, KLP, and PSA, while VRE maintained a nearly constant PPV across all four models.

We next perform model-to-model comparisons to isolate the relative strengths of EHR and AR across species. Regarding sensitivity, the addition of EHR to either the standalone MALDI-TOF or the joint MALDI+AR model did not yield consistent gains across species (Figure 2b,c). However, the tri-modal approach (MALDI+AR+EHR) outperformed both its bi-modal counterparts for CB, PRV, and SER (Figure 2d), suggesting a degree of complementarity between EHR and AR signals in these specific contexts.

Nonetheless, the impact on PPV was more pronounced. While the effect of adding AR to the MALDI+EHR model was heterogeneous across species (Figure 2g), the integration of EHR with both MALDI and MALDI+AR resulted in consistent PPV gains (Figure 2e,f). Notably, high-complexity species characterized by a large number of outbreak clusters (e.g., PSA, PR, and MRSA) demonstrated marked improvements in PPV through EHR integration. This finding addresses a known bottleneck: our previous work observed that both MALDI-TOF and AR are often insufficient to elevate PPV for these complex pathogens[17]. In summary, these results suggest that (1) multimodal integration enhances performance in a species-specific manner, (2) EHR and AR data exhibit complementarity in certain contexts, and (3) EHR integration facilitates performance gains for species with high cluster complexity.

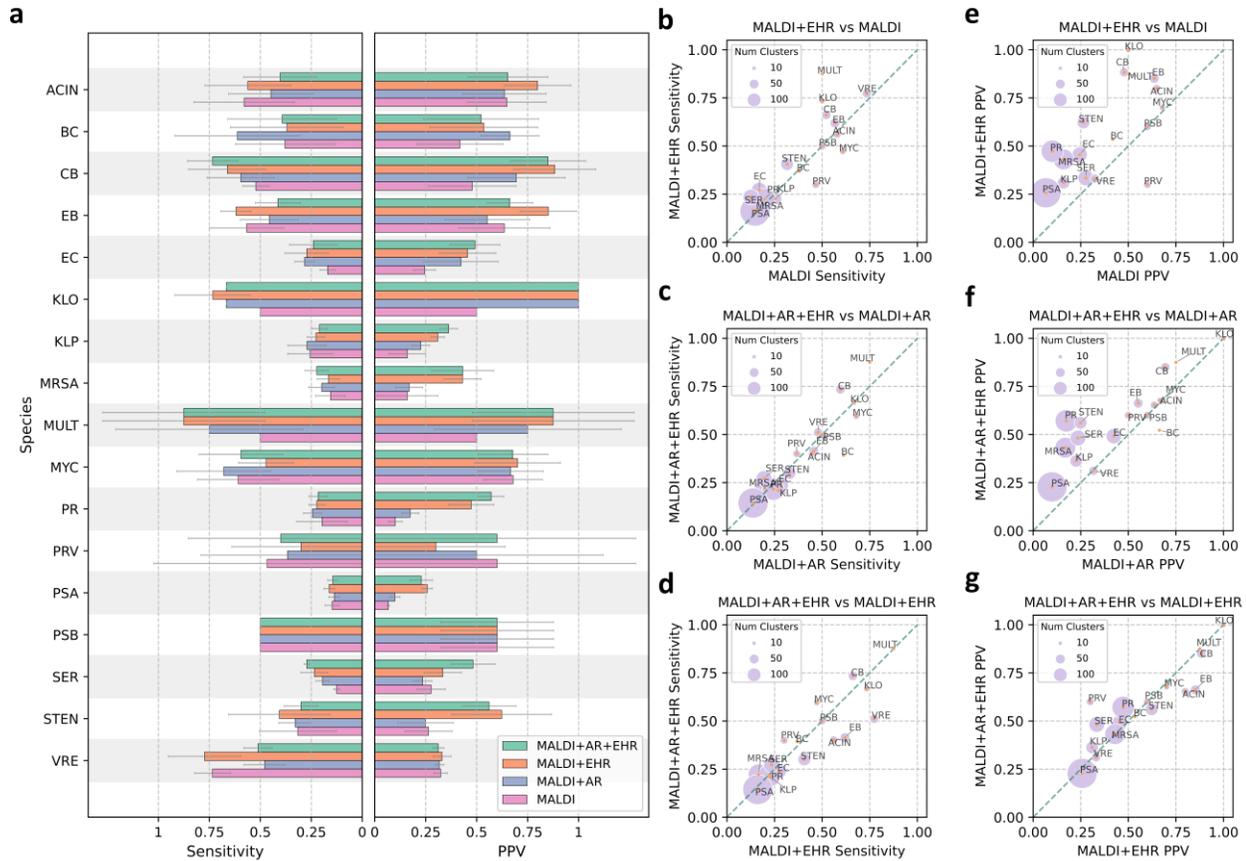

**Figure 2.** Per-species outbreak detection performance. **a**. Grouped bar plots for sensitivity (left) and PPV (right). **b**–**g**. Scatterplots comparing sensitivity (**b**–**d**) and PPV (**e**–**g**) across model pairs (MALDI, MALDI+EHR, and MALDI+AR+EHR). Circle size reflects the number of ground-truth outbreak clusters. Points above the diagonal indicate performance improvements following modality integration.

*Tiered surveillance reduces WGS dependency*: To assess the potential of the alternative modalities in reducing reliance on WGS, we tested our proposed tiered surveillance paradigm. Under this framework, isolates not assigned to a predicted cluster by the alternative modality are triaged for WGS confirmation. For each model, we evaluated 20 equally spaced distance thresholds across the Euclidean distances in the training data representations, and calculated the percentage of isolates needing WGS alongside the resulting combined sensitivity and combined PPV.

The relative utility of these modalities is reflected in the position of their performance curves. Specifically, for a fixed percentage of isolates referred to WGS sequencing, the modalities yielding higher combined sensitivity and PPV demonstrate higher outbreak detection utility. Conversely, to achieve a target level of combined sensitivity or PPV, a superior modality is one that requires a smaller percentage of isolates to be sent for WGS. Note that in practice, clinical teams can select an operational threshold along these curves based on specific institutional budgets and performance requirements.

We use *Escherichia coli* (EC) and *Proteus mirabilis* (PR) as examples to illustrate the efficacy of this tiered framework. Across all models, the curves for combined sensitivity (Figure 3a,c) and PPV (Figure 3b,d) versus the percentage of isolates requiring WGS consistently lie above the diagonal null baseline. This lift confirms that MALDI-TOF, AR, and EHR provide informative signals under the tiered surveillance paradigm. Furthermore, the integration of EHR and AR with MALDI-TOF data yields higher performance curves, with EHR integration providing the most substantial gains in utility. In summary, the tiered surveillance paradigm effectively reduces the burden on genomic sequencing by leveraging the informative signals provided by integrated clinical and microbial data.

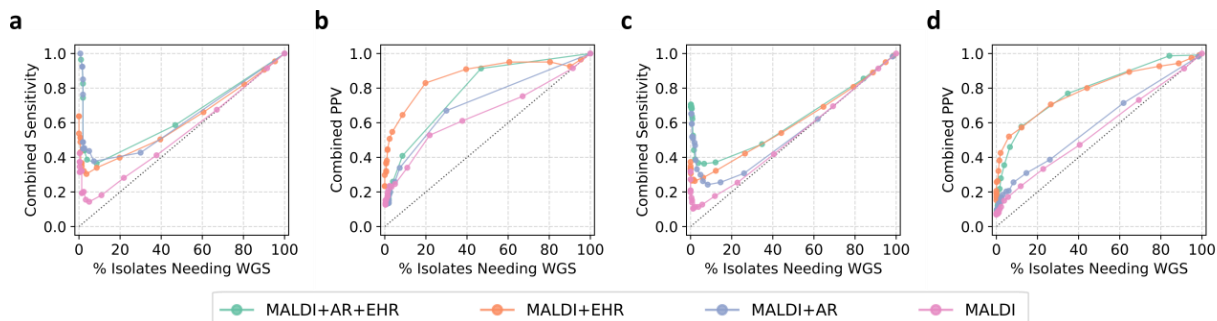

**Figure 3.** Evaluation of the tiered surveillance paradigm. We plot the combined sensitivity and PPV as a function of the percentage of isolates requiring WGS referral for EC (**a, b**) and PR (**c, d**). The diagonal line represents a null baseline that corresponds to a model whose clustering provides no information.

**Table 3**. Risk statistics of the top 15 and bottom 3 procedure groups ranked by descending **Mean Risk**. **Mean** and **Max Risk** represent the average and maximum Wasserstein distance across all affected clusters, respectively, between the distributions of predicted intra-cluster SNP distances before and after the removal of a procedure group. **Mean Δ SNP$_{pred}$** indicates the average change in mean predicted intra-cluster SNP distances. **# Isolates** denotes the count of isolates from patients who underwent a procedure within the specified group. **Cluster-wise statistics** highlight the three clusters with the highest patient-to-cluster size ratios, reported in the format: "cluster size (number of isolates in cluster, cluster-specific risk score)".

| Group Description | Mean Risk | Max Risk | Mean Δ SNP$_{pred}$ | # Isolates | Cluster-wise statistics |
|---|---|---|---|---|---|
| GUIDEWIRE URETER | 6.976 | 8.718 | 6.976 | 4 | 5(3,8.341), 22(1,2.881) |
| DOCUSATE | 6.011 | 12.154 | 5.852 | 14 | 5(5,10.908), 5(1,2.746), 5(1,1.962) |
| GASTROSTOMY | 5.927 | 11.098 | 5.891 | 13 | 6(5,8.936), 5(1,3.672), 5(1,3.636) |
| Esophagoscopy | 5.177 | 17.918 | 4.999 | 57 | 5(5,15.036), 7(7,5.035), 11(7,6.995) |
| Venography | 4.948 | 6.745 | 4.046 | 3 | 6(2,6.582), 22(1,1.680) |
| D5 | 4.903 | 9.326 | 4.749 | 9 | 5(2,6.886), 6(2,4.071), 5(1,3.170) |
| Endoscopy | 4.810 | 7.673 | 4.756 | 9 | 7(3,6.345), 7(2,7.150), 5(1,4.886) |
| AQUAPHOR | 4.796 | 7.456 | 4.775 | 7 | 5(1,4.648), 5(1,4.946), 11(2,7.411) |
| TRAY CATH URET | 4.613 | 9.153 | 4.576 | 18 | 6(3,8.976), 6(2,4.286), 14(3,4.931) |
| Stent | 4.558 | 9.153 | 4.494 | 33 | 5(3,8.341), 6(3,8.976), 5(2,4.118) |
| Surgical Sealer | 4.549 | 9.153 | 4.488 | 25 | 6(6,7.629), 5(3,4.911), 6(2,3.464) |
| Venipuncture | 4.549 | 17.918 | 4.260 | 119 | 9(9,5.162), 6(5,4.920), 5(4,17.244) |
| Surgical Staple | 4.428 | 16.439 | 4.180 | 32 | 5(3,4.911), 7(4,8.744), 6(3,3.951) |
| NOREPINEPHRINE | 4.344 | 10.228 | 4.200 | 17 | 6(3,2.564), 5(2,8.718), 6(2,10.228) |
| Arterial puncture | 4.335 | 8.609 | 4.074 | 23 | 11(6,7.709), 6(3,2.758), 11(3,3.915) |
| ... | | | | | |
| Magnetoencephalography | 1.423 | 1.423 | 0.969 | 1 | 46(1,1.423) |
| LARYNGOSCOPY | 1.254 | 1.790 | 1.254 | 2 | 6(1,0.717), 6(1,1.790) |
| US URINE CAPACITY MEASURE | 1.242 | 1.242 | 0.341 | 1 | 46(1,1.242) |

*Interrogating EHR feature vectors reveals potentially high-risk procedures*: Following the protocol described in our methodology, we retrospectively analyzed the EHR data representations to identify procedures that may indicate high-contamination-risk routes. To ensure the numerical stability of the Wasserstein distance calculation, we restricted this risk score analysis to clusters containing at least five isolates.

Evaluation of procedures with the top 15 risk scores (Table 3) highlights two types of transmission profiles: equipment vulnerabilities and workflow exposures. First, invasive interventions utilizing reusable equipment, such as Ureter Guidewires (mean risk: 6.98) and Esophagoscopy (max risk: 17.92), exhibit the highest overall risk. The potential contamination of such equipment facilitates localized clusters where 100% of patients shared the same procedure (e.g., 7/7 patients in cluster for Esophagoscopy). Second, routine and high-frequency tasks point to workflow exposures. For example, Venipuncture affects the largest absolute number of patients (119 isolates, 9/9 patients in largest cluster), while routine oral medication administration (5/5 patients in cluster for Docusate) highlights transmission risks embedded in standard daily care. This distribution indicates exposures linked to shared clinical environments, multi-use medication carts, or the inherent challenges of mitigating transmission during high-turnover care. Collectively, these EHR-derived insights provide medical institutions with actionable targets for risk mitigation and screening to enforce best sanitizon practices: optimizing reprocessing protocols for medical equipment while systematically implementing environmental control measures in high-volume care spaces.

**Discussion**
In this study, we develop and evaluate a multimodal machine learning framework for hospital outbreak detection that integrates MALDI-TOF mass spectrometry, antimicrobial resistance (AR) patterns, and electronic health records (EHR) data. These standard clinical modalities require less capital and fewer resources than the current gold standard of whole genome sequencing (WGS). Our analysis shows that these modalities encode mutually complementary information that collectively improves detection performance, though benefits vary by metric and pathogen. Notably, while MALDI-TOF provides a strong baseline, adding EHR data improves positive predictive value (PPV) for species with complex cluster structures, where microbial signals alone are insufficient. These results support the utility of multimodal, integrated surveillance architectures.

Rather than replacing WGS, we propose a tiered surveillance paradigm. Isolates are initially screened using the multimodal framework, reserving WGS confirmation for unclustered cases. When WGS is available, this tiered approach offers flexibility to balance costs with required performance thresholds. However, to realize this paradigm in operational settings, future research must conduct detailed economic analyses regarding the costs of these modalities and the impact of their latency on infection control.

Additionally, using retrospective analysis of data via counterfactual feature ablation of the EHR representations provides a way to identify facility-specific vulnerabilities and offers some level of epidemiological interpretability. This type of analysis can reveal potentially high-risk transmission routes, in our example associated with invasive medical equipment or high-turnover care environments. By identifying such vulnerabilities, our approach can provide infection prevention teams with targeted insights to purposively review the standards of practice and to focus proactive mitigation of risk campaigns.

Future work will focus on improving the framework's overall sensitivity to close the performance gap with WGS. We plan to explore advanced architectures, such as transformer-based models, and incorporate spatial metadata (e.g., patient room assignments) into the EHR feature space to boost discriminative power. We also plan to conduct deeper investigations into the underlying population genetics and resistance biology of the evaluated pathogens, which will help elucidate why MALDI-TOF, AR, and EHR each perform well in the specific microbial contexts.

**Acknowledgements**
This work was funded in part by the National Institute of Allergy and Infectious Diseases, National Institutes of Health (NIH) grant R01AI127472 and by the National Science Foundation award 2406231